\documentclass{article}

\usepackage[final]{nips_2017}

\usepackage[utf8]{inputenc} 
\usepackage[T1]{fontenc}    
\usepackage{hyperref}       
\usepackage{url}            
\usepackage{booktabs}       
\usepackage{amsfonts}       
\usepackage{nicefrac}       
\usepackage{microtype}      
\usepackage{graphicx,graphics}
\usepackage{enumitem, kantlipsum}
\usepackage[utf8]{inputenc}
\usepackage{amssymb}
\usepackage{amsmath}
\usepackage{physics}
\usepackage{verbatim}
\usepackage{hyperref}
\usepackage{bm}
\usepackage{multirow}
\usepackage{tablefootnote}
\usepackage{hyperref}

\usepackage[numbers]{natbib}
\title{Regularization approaches for support vector machines with applications to biomedical data}
\author{
  Daniel Lopez-Martinez \\
  Harvard-MIT Health Sciences and Technology \\
  Massachusetts Institute of Technology\\
  \texttt{dlmocdm@mit.edu}
}

\begin{document}
\maketitle

\begin{abstract}
The support vector machine (SVM) is a widely used  machine learning tool for classification based on statistical learning theory. Given a set of training data, the SVM finds a hyperplane that separates two different classes of data points by the largest distance. While the standard form of SVM uses $L_2$-norm regularization, other  regularization approaches are particularly attractive for biomedical datasets where, for example, sparsity and interpretability of the classifier's coefficient values are highly desired features. Therefore, in this paper we consider different types of regularization approaches for SVMs, and explore them in both synthetic and real biomedical datasets. 


\end{abstract}


\section{Introduction}

A support vector machine (SVM) is a supervised learning discriminative binary classifier formally defined by a separating hyperplane. In other words, given a set of labeled training data $(\textbf{x}_1,y_1),...,(\textbf{x}_n,y_n)$ where $\textbf{x} \in \mathbb{R}^p$ is a vector with $p$ predictor variables and $y\in \{-1,+1 \}$ denotes the class label, the algorithm outputs an optimal hyperplane $f(\textbf{x}) = \bm{\beta}^\intercal \textbf{x} + \beta_0$ which categorizes new examples according to the decision rule $\Phi (\textbf{x}) = \textnormal{sign}(f(\textbf{x}))$. The hyperplane is chosen in a way that separates the two classes of data points by the largest distance. Solving such a problem is an exercise in convex optimization. In the case of separable data, the popular setup (known as primal form) is given by: 
\begin{equation} \label{eq:primal-separable}
\min_{\beta_0,\bm{\beta}} \frac{1}{2} ||\bm{\beta}||^2_2, \quad \textnormal{subject to  } \; y_i (\beta_0 + \bm{x}_i^\intercal \bm{\beta}) \ge 1 \quad (\forall i)
\end{equation}
A bit of linear algebra shows that $\frac{1}{||\bm{\beta}||} (\beta_0 + \bm{x}_i^\intercal \bm{\beta}) $ is the signed distance from $\bm{x}_i$ to the decision boundary. When the data is not separable, the criterion is modified to:
\begin{equation} \label{eq:primal-nonseparable}
\min_{\beta_0,\bm{\beta},\bm{\xi}} \frac{1}{2}||\bm{\beta}||^2_2 + C \sum_{i=1}^n \xi_i, \quad \textnormal{subject to  } \ y_i (\beta_0 + \bm{x}_i^\intercal \bm{\beta}) \ge 1 -\xi_i, \; \xi_i \ge 0 \quad  (\forall i)
\end{equation}
where $\xi_i$ are the non-negative slack variables that allow points to be on the wrong side of their soft margin as well as the decision boundary, therefore describing the overlap between the two classes, and $C$ is a cost parameter that controls the amount of overlap. Note that by solving for the Lagrangian dual of the above problem, one can obtain a simplified problem that is efficiently solvable by quadratic programming algorithms, known as the dual form of the SVM.

Alternatively, we can formulate the above problem (eq. \ref{eq:primal-nonseparable}) in the regularization framework, using a \textit{loss} + \textit{penalty} criterion:
\begin{equation} \label{eq:errorregularizer}
\min_{\beta_0, \bm{\beta}} \ \underbrace{\sum_{i=1}^n [1-y_i (\beta_0 + \bm{x}_i^\intercal \bm{\beta})]_+}_\text{loss} + \underbrace{\frac{\lambda}{2} || \bm{\beta}||^2_2}_\text{regularizer}
\end{equation} 
where $\lambda$ is the regularization parameter (it relates to $1/C$ in eq. \ref{eq:primal-nonseparable}) , and the function $(1- a)_+ = \max(0, 1- a)$ is the hinge loss. Equation \ref{eq:errorregularizer}  also admits a class of more flexible, nonlinear generalizations:
\begin{equation} \label{eq:generalization}
\min_{f \in \mathcal{H}}\sum_{i=1}^n V(y_i, f(\bm{x}_i)) + \lambda J(f)
\end{equation}
where $V(y_i, f(\bm{x}_i))$ is the loss function (e.g. the hinge loss),  $f(\bm{x})$ is an arbitrary function in some Hilbert space $\mathcal{H}$, and $J(f)$ is a functional that measures the "roughness" of $f$ in $\mathcal{H}$. Under this generalization, the extension to nonlinear kernel SVMs becomes straightforward. In that case, $f(\bm{x}) = \beta_0 + g(\bm{x})$ , and $J(f)=J(g)$ is a norm in a RKHS of functions $\mathcal{H}_K$ generated by a positive definite kernel $K(x,x')$. Consequently, $f(\bm{x}) = \beta_0  + \sum_{i=1}^n \theta_i K(\bm{x},\bm{x}_i)$, and using the hingle loss eq. \ref{eq:generalization} reduces to:
\begingroup\makeatletter\def\f@size{9}\check@mathfonts
\def\maketag@@@#1{\hbox{\m@th\large\normalfont#1}}%
\begin{equation} \label{eq:generalizationkernelform}
\min_{\beta_0, \bm{\theta}} \ \sum_{i=1}^n \left [1-y_i \left (\beta_0 + \sum_{i=1}^n \theta_i K(\bm{x},\bm{x}_i) \right )\right ]_+ 
+ \frac{\lambda}{2} \sum_{j=1}^{n} \sum_{j'=1}^{n} \theta_j \theta_{j'} K(\bm{x}_j,\bm{x}_{j'})
\end{equation}
\endgroup


\section{SVM regularization approaches }

In the previous section, we employed the $L_2$-norm penalty in equations \ref{eq:primal-separable}, \ref{eq:primal-nonseparable} and \ref{eq:errorregularizer}.
By shrinking the magnitude of the coefficients, the $L_2$-norm penalty reduces the variance of the estimated coefficients, and thus can achieve better prediction accuracy. However, the $L_2$-norm penalty cannot produce sparse coefficients and hence cannot automatically perform variable selection. This is a major limitation for applying SVM to do classification in some high-dimensional data where variable selection is essential for providing reasonable interpretations. This is usually the case in biomedical data, where obtaining a good classifier does not suffice – we also want to know which variables are relevant to our classification problem (e.g. which genes in a gene panel can be used for disease diagnosis). Therefore, we expand the traditional $L_2$-norm SVM formulation to include sparsity and variable selection, using different regularization approaches and testing them in real biomedical datasets. For simplicity, we consider the linear kernel, but our results can easily be extended to other kernels.  
 
\subsection{L1-norm regularization} \label{sec:l1lasso}
An alternative to using the $L_2$-norm penalty in eq. \ref{eq:errorregularizer} is to use the $L_1$-norm of $\bm{\beta}$, i.e. the \textit{lasso} penalty:
\begin{equation} \label{eq:l1}
\min_{\beta, \bm{\beta}} \sum_{i=1}^n [1-y_i (\beta_0 + \bm{x}_i^\intercal \bm{\beta})]_+ + \lambda || \bm{\beta}||_1
\end{equation}
Several algorithms exist to solve the above problem. For example, it can be expressed in primal form and solved using a standard linear programming software package (e.g. \cite{1norm}), or it can be solved with a coordinate descent algorithm (\cite{coordinatedescent,liblinear} for the squared hinge loss). 
Similar to the $L_2$-norm penalty, the $L_1$ norm also shrinks the fitted coefficients toward zero, hence also reducing the variance of the fitted
coefficients. However, in this case, owing to the mathematical properties of the $L_1$ norm (e.g. nondifferentiability at 0), making $\lambda$ sufficiently large will cause some of the fitted coefficients be exactly zero. Therefore, the lasso penalty promotes sparsity and hence it does a kind of feature selection, which is not the case for the $L_2$ penalty. 

Unfortunately, the $L_1$-norm penalty suffers from an important disadvantage in terms of feature selection: when there are several highly correlated input variables in the data set, and they are all relevant to the output variable, the lasso penalty will tend to pick one or few of them and shrink the rest to zero. In other words, the $L_1$ norm fails to do "grouped selection". This is particularly important in the context of biomedical datasets. For example, let's consider a gene panel for the diagnosis of a multigenic disease that requires abnormally high expression levels of all genes in a particular set of genes regulating one or several biological pathways. An ideal classifier should be able to automatically include the whole group of relevant genes. However, because the correlation among these genes will be high, the $L_1$-norm will select only one gene from the set of relevant genes, and set the corresponding coefficients for the other relevant genes to zero. 
One  way to overcome this limitation is to apply the elastic net penalty to the SVM, as discussed in section \ref{sec:elasticnet}.

Despite this disadvantage, the $L_1$-norm penalty also presents one major advantage, that is, it's extension to multi-class classification problems is straightforward. This is specially advantageous since SVMs are inherently two-class classifiers. The traditional way to do $k$-class classification with SVMs is to use one of the \textit{one-vs-all} approach: for each class $c=1,...,k$, we build the $c$th classifier such that we let the positive examples be all the points in class $c$, and we let the negative examples be all the points not in class $c$. Let $f_c$ be the $c$th classifier. Then, the decision rule is simply $f(\bm{x}) = \text{argmax}_c f_c (\bm{x})$. Unfortunately, this method has multiple disadvantages. For example, when the number of classes becomes large, each binary classification becomes highly unbalanced, with a small fraction of instances in one class. In the case of non-separable classes, the class with smaller fraction of instances tends to be ignored, leading to degrading performances. In addition to this, from the perspective of feature selection, one feature is relevant for all classes if it is selected in one binary classification. In the presence of many irrelevant features, this usually results in more than necessary features and therefore it has an adverse effect on both classification and interpretability. 

Alternatively, instead of combining separately trained binary SVM classifiers, a single optimization problem can be formulated to consider all classes and be solved at once, as done by Crammer and Singer \cite{crammersinger} for the $L_2$-norm regularization case. Here, we propose and solve an alternative \textit{all-in-one}  multi-class optimization problem for $L_1$-norm regularization based on \cite{L1MSVM}. The first step is to generalize the binary hinge loss for the multi-class case. Several such generalizations of the hinge loss exist in the literature, such as $V(f,z_i)=\sum_{c \ne y_i} [1-(f_{y_i}(x_i))-(f_{c}(x_i))]_+$ \cite{vapnik98}, $V(f,z_i)=\sum_{c\ne y_i} [f_c (x_i) +1]_+$ \cite{lee04} or $V(f,z_i)=[1-\min_{c,c\ne y_i} ( f_{y_i}(x_i) - f_c (x_i) )]_+$ \cite{liu06}, where $z_i=(x_i,y_i)$. Here, we use the later, as done in the $L_2$ multi-class SVM \cite{crammersinger}. Then, the multi-class SVM with $L_1$ penalty can be expressed in primal form as:
\begin{equation} \label{eq:l1msvm-primal}
\begin{split}
& \min_{\bm{\beta},\beta_0,\bm{\xi}} \; \sum_{i=1}^n\xi_i + \lambda \sum_{c=1}^{k} \sum_{i=1}^{p} |\beta_{c,j}| \\ 
&  \textnormal{subject to } f_{yi} - \max_{k \ne yi} f_{k} (x_i) \ge 1 - \xi_i,\quad \xi_i \ge 0, \; \forall i
\end{split}
\end{equation} 
where $k$ is the number of classes under consideration. Liu and Shen \cite{liu06} show that this formulation has the natural interpretation of minimizing $V(f,z_i)=[1-\min_{c,c\ne y_i} ( f_{y_i}(x_i) - f_c (x_i) )]_+$. Then, eq. \ref{eq:l1msvm-primal} can be expressed as a linear programming optimization problem:
\begingroup\makeatletter\def\f@size{9}\check@mathfonts
\def\maketag@@@#1{\hbox{\m@th\large\normalfont#1}}%
\begin{equation} 
\begin{split}
 \min_{\bm{\beta},\beta_0,\bm{\xi}} \; & \sum_{i=1}^n\xi_i + \lambda \sum_{c=1}^{k} \sum_{i=1}^{p} (\beta_{c,j}^+ -  \beta_{c,j}^-) \\
\textnormal{s.t.: } \; & 
\sum_{j=1}^p (\beta_{y,j}^+ - \beta_{y,j}^+)x_{i,j} - \sum_{j=1}^p  (\beta_{c,j}^+ - \beta_{c,j}^+)x_{i,j} \ge 1 - \xi_i, \: \textnormal{for } i=1,...,n; \; c=1,...,k; \; c \ne y_i 
\\
& \sum_{c=1}^k \beta_{k,0} = 0; \ \sum_{c=1}^k (\beta_{c,j}^+ - \beta_{c,j}^-) =0, \ j=1,...,p \\
& \beta_{c,j}^+ \ge 0, \ \beta_{c,j}^- \ge 0, \ \xi_i \ge 0, \ \forall c,j,i
\end{split}
\end{equation} \endgroup
where $|\beta_{c,k}| = \beta_{c,j}^+ - \beta_{c,j}^-$; $\beta_{c,j}^+ = \beta_{c,j}$ if $\beta_{c,j}\ge 0$ or 0 otherwise; $\beta_{c,j}^- = -\beta_{c,j}$ if $\beta_{c,j} \le 0$ or 0 otherwise. These changes are necessary to eliminate the absolute value operation, so that eq. \ref{eq:l1msvm-primal} can be solved by a linear programming software package \cite{algo}.

This multi-class SVM optimization problem results in $k$ linear decision functions $f_c(\bm{x}) = \beta_0 + \bm{x}_i^\intercal \bm{\beta}_c$ with $\bm{\beta}_c \in \mathbb{R}^p$. The decision rule is simply $f(\bm{x}) = \text{argmax}_c f_c (\bm{x})$. 

\subsection{Elastic net regularization} \label{sec:elasticnet}

The elastic-net penalty is a mixture of the $L_1$-norm penalty and the $L_2$-norm penalty. 
\begin{equation} \label{eq:elasticnet}
\min_{\beta, \bm{\beta}} \sum_{i=1}^n [1-y_i (\beta_0 + \bm{x}_i^\intercal \bm{\beta})]_+ + \frac{\lambda_2}{2} ||\bm{\beta}||_2^2 + \lambda_1 || \bm{\beta}||_1
\end{equation}
where both $\lambda_1$ and $\lambda_2$ are the regularization parameters.  
Here, the role of the $L_1$-norm penalty is to allow variable selection, whereas the role of the $L_2$-norm penalty is to help groups of correlated variables get selected together, that is, highly correlated input variables will have similar fitted coefficients \cite{doubly}. This is called the grouping effect, and it is a highly desirable feature in biomedical applications: in many cases, the presence of a disease will be characterized by an increase or decrease in more than one variable, which will likely be highly correlated. Therefore, due to this grouping, the elastic net will select all the correlated variables together, therefore improving the interpretability of the nonzero classifier coefficients.

To show the grouping property of the elastic net \cite{doubly} with hinge loss $V(y,f(\bm{x}))= [1-yf(\bm{x})]_+$, let's first note that  the hinge loss is  Lipschitz continuous with constant $M=1$, that is, $|V(t_1)-V(t_2)| \le M |t_1 - t_2|$. 
Then, let $\hat{\beta}_0$ and $\hat{\bm{\beta}}$ be the solution to  eq. \ref{eq:elasticnet}, and consider another set of coefficients $\hat{\beta}_0^*$ and $\hat{\bm{\beta}}^*$ such that:
\begin{equation} 
\hat{\beta}_0^*=\hat{\beta}_0, \qquad 
\hat{{\beta}}_{j'}^*=\begin{cases}
	\frac{1}{2} (\hat{\beta}_{j}+\hat{\beta}_{l}), & \quad \textnormal{if} \; j'=j \textnormal{ or } j'=l \\
    \hat{{\beta}}_{j} & \quad \textnormal{otherwise.}
\end{cases}
\end{equation}
Clearly, since $(\hat{\beta}_0, \hat{\bm{\beta}})$ is the solution to  eq. \ref{eq:elasticnet}, we have: 
\begin{equation} \label{eq:inequality}
\left ( \sum_{i=1}^n V(y_i,\hat{\beta}_0^* + \bm{x}_i^\intercal \hat{\bm{\beta}}^*) + \frac{\lambda_2}{2} ||\hat{\bm{\beta}}^*||_2^2 + \lambda_1 ||\hat{\bm{\beta}}^*||_1  \right ) -
\left ( \sum_{i=1}^n V(y_i,\hat{\beta}_0 + \bm{x}_i^\intercal \hat{\bm{\beta}}) + \frac{\lambda_2}{2} ||\hat{\bm{\beta}}||_2^2 + \lambda_1 ||\hat{\bm{\beta}}||_1 \right )
\ge 0
\end{equation}
Now, let's note that:
\begin{equation} \label{eq:inequality2}
\begin{split} 
\sum_{i=1}^n \left [V(y_i, \hat{\beta}_0^* + \bm{x}_i^\intercal \hat{\bm{\beta}}^*)-V(y_i, \hat{\beta}_0 + \bm{x}_i^\intercal \hat{\bm{\beta }}) \right ]
\le \sum_{i=1}^n \left |V(y_i, \hat{\beta}_0^* + \bm{x}_i^\intercal \hat{\bm{\beta}}^*)-V(y_i, \hat{\beta}_0 + \bm{x}_i^\intercal \hat{\bm{\beta }}) \right |
\end{split}
\end{equation}
Using Lipschitz continuity, we obtain:
\begingroup\makeatletter\def\f@size{9}\check@mathfonts
\def\maketag@@@#1{\hbox{\m@th\large\normalfont#1}}%
\begin{equation} \label{eq:proof}
\begin{split}
\sum_{i=1}^n \left |V(y_i, \hat{\beta}_0^* + \bm{x}_i^\intercal \hat{\bm{\beta}}^*)-V(y_i, \hat{\beta}_0 + \bm{x}_i^\intercal \hat{\bm{\beta }}) \right | & 
\le \sum_{i=1}^n M  \left |  y_i (\hat{\beta}_0^* + \bm{x}_i^\intercal \hat{\bm{\beta}}^*) -    y_i (\hat{\beta}_0 + \bm{x}_i^\intercal \hat{\bm{\beta}})            \right | \\
& = \sum_{i=1}^n M \left | \bm{x}_i^\intercal (\hat{\bm{\beta}}  - \hat{\bm{\beta}} ) \right | \\
& =  \sum_{i=1}^n M \left | \frac{1}{2} (x_{i,j} - x_{i,l}) (\hat{{\beta}}_j  - \hat{{\beta}}_l ) \right |  \\
& = \frac{M}{2}  \left |  \hat{{\beta}}_j  - \hat{{\beta}}_l \right | \sum_{i=1}^n |x_{i,j} - x_{i,l}|
= \frac{M}{2}  \left |  \hat{{\beta}}_j  - \hat{{\beta}}_l \right | \cdot  ||\bm{x}_j-\bm{x}_l||_1
\end{split}
\end{equation}
\endgroup
In addition to this, we have: 
\begingroup\makeatletter\def\f@size{9}\check@mathfonts
\def\maketag@@@#1{\hbox{\m@th\large\normalfont#1}}%
\begin{equation} \label{eq:additional}
\begin{split}
& ||\hat{\bm{\beta}}^*||_1 - ||\hat{\bm{\beta}}||_1  = |\hat{\beta}_j^*| + |\hat{\beta}_l^*| - |\hat{\beta}_j| - |\hat{\beta}_l|
= |\hat{\beta}_j + \hat{\beta}_l| - |\hat{\beta}_j| - |\hat{\beta}_l| \le 0 \\
& ||\hat{\bm{\beta}}^*||_2^2 - ||\hat{\bm{\beta}}||_2^2   = |\hat{\beta}_j^*|^2 + |\hat{\beta}_l^*|^2 - |\hat{\beta}_j|^2 - |\hat{\beta}_l|^2 = -\frac{1}{2} |\hat{\beta}_j-\hat{\beta}_l|^2
\end{split}
\end{equation}
\endgroup

Combining equations \ref{eq:inequality}, \ref{eq:inequality2},\ref{eq:proof} and \ref{eq:additional}, we obtain 
\begingroup\makeatletter\def\f@size{9}\check@mathfonts
\def\maketag@@@#1{\hbox{\m@th\large\normalfont#1}}%
\begin{equation} \label{eq:groupingeq1}
\frac{M}{2}  \left |  \hat{{\beta}}_j  - \hat{{\beta}}_l \right | \cdot  ||\bm{x}_j-\bm{x}_l||_1  
- \frac{\lambda_2}{2} \left |  \hat{{\beta}}_j  - \hat{{\beta}}_l \right |^2 \ge 0
\end{equation}
\endgroup
When the input variables $\bm{x}_j$ and $\bm{x}_l$ are centered and normalized, we have that $||\bm{x}_j - \bm{x}_l ||_1 \le \sqrt{n} \sqrt{||\bm{x}_j - \bm{x}_l||^2_2} = \sqrt{n} \sqrt{2(1-\rho)}$, where $\rho=\textnormal{cor}(\bm{x}_j,\bm{x}_l)$ is the correlation between $\bm{x}_j,\bm{x}_l$ . Then, and eq.  \ref{eq:groupingeq1} can be simplified to
\begin{equation} \label{eq:groupingeq2}
|\hat{\beta}_j-\hat{\beta}_l| \le \frac{\sqrt{n} M}{\lambda_2} \sqrt{2(1-\rho)}
\end{equation}
where $M=1$ for the hinge loss. This demonstrates the grouping property of the elastic net. Note also that eq. \ref{eq:groupingeq2} holds for all $\lambda_1 \ge 0$. Therefore, the grouping effect is due to the $L_2$-norm penalty.

Finally, note that there exist multiple algorithmic implementations that solve the elastic net SVM. For example, Zhou et al. \cite{elasticnetgpu} have developed a parallel solver  that  utilizes GPUs and multi-core CPUs.

\subsection{k-support norm regularization}

Another regularization approach is based on the $k$-support norm, proposed by \cite{ksupport}. This is a sparsity regularization method that balances the $L_1$ and $L_2$ norms over a linear function, very similar to the elastic net. The $k$-support norm is based on the convex hull of $S_k^{(2)} := \{ \bm{\beta} \: | \: ||\bm{\beta}||_0 \le k, ||\bm{\beta}||_2 \le 1\}$, that is, $C_k := \textnormal{conv}(S_k^{(2)})$. For $k>1$, the $k$-support norm is a tighter convex relaxation than the elastic net, and is therefore a better convex constraint than the elastic net when seeking a sparse, low $L_2$-norm linear predictor that combines the uniform shrinkage of an $L_2$ penalty for the largest components, with the sparse shrinkage of an $L_1$ penalty for the smallest components.

The $k$-support norm can be computed as:
\begingroup\makeatletter\def\f@size{9}\check@mathfonts
\def\maketag@@@#1{\hbox{\m@th\large\normalfont#1}}%
$$ 
||\bm{\beta}||_k^{sp}= \left (  \sum^{k-r-1}_{i=1} (|\bm{\beta}|_i^\downarrow)^2  +\frac{1}{r+1} \left( \sum_{i=k-r}^{d} |\bm{\beta} |_i^\downarrow \right)^2 \right )^\frac{1}{2}
$$ \endgroup
where $|\beta|_i^\downarrow$ is the $i$th largest element of the vector $\bm{\beta}$ and $r$ is a unique integer in $\{0,...,k-1 \}$ satisfying $|\bm{\beta}|_{k-r-1}^\downarrow > \frac{1}{r+1} \sum_{i=k-r}^{d} |\bm{\beta}|_i^\downarrow \ge |\bm{\beta}|_{k=r}^\downarrow $

The $k$-support normalization leads to the following primal form SVM optimization problem: 
\begin{equation}
\begin{split}
& \min_{\bm{\beta},\bm{\xi}} \lambda ||\bm{\beta}||_k^{sp} + \sum_{i=1}^n \xi_i \\
& \textnormal{s.t.} y_i (\beta_0 + \bm{x}_i^\intercal \bm{\beta}) \ge 1 -\xi_i, \xi_i \ge 0, \forall i
\end{split}
\end{equation}
where $\lambda>0$ is the regularization parameter, and $k\in\{ 1,...,p \}$, where $p$ is the dimension of the feature space. Note that $k$ negatively correlates with sparsity.

This algorithm also uses the hinge loss, as in our previous examples, and leads to sparse but correlated subsets of selected features. It has been implemented \footnote{\url{https://github.com/gkirtzou/ksup_svm }} using the Nesterov's accelerated gradient descent algorithm \cite{kimplementation}.



\section{Empirical comparisons and discussion}

Here, we study the sparsity and grouping effect properties of the different regularization approaches considered, in both synthetic and real biomedical datasets.

\subsection{Sparsity  and grouping effect: L2, L1, elastic net, k-support}

\paragraph{Sparsity analysis} We demonstrate the different sparsity properties of a  the above regularization approaches on the Wisconsin breast cancer dataset \footnote{\url{https://archive.ics.uci.edu/ml/datasets/Breast+Cancer+Wisconsin+(Diagnostic)}} containing a total of 569 instances of 30 features  computed from digitized images of  fine needle aspirates  of breast mass, with labels $y_+ = 1 $ for malignant cancer and $y_- = -1$ for benign tumors. The dataset was divided in a training set (66\%) and a test set (33\%). We performed 10-fold cross-validation on the training set to optimize the hyperparameters of the SVM classifier. The results, shown in Table \ref{tab:breastcancer}, show the average results for 10 repetitions and   indicate that the $k$-support SVM leads to the best performance on this dataset (93.57\%),  followed by the elastic net SVM (83.07\%). Both approaches lead to sparse coefficients, with 9.2 and 8.6 non-zero coefficients respectively, out of  30 coefficients. In addition to this, Table \ref{tab:breastcancer} also shows the results for the same dataset when "contaminated" with 30 additional noise features drawn from a N(0,1) distribution. In this case, the number of relevant non-zero coefficients (the ones corresponding to the original non-noise features) stays the same for all algorithms, but the $L_1$ and $L_2$ regularizations select considerably more non-relevant features. In all cases, the $k$-support and elastic net SVMs lead to both the best performances and higher sparsity.

\begin{figure}[t!] 
	\centering
	\includegraphics[width=1\textwidth]{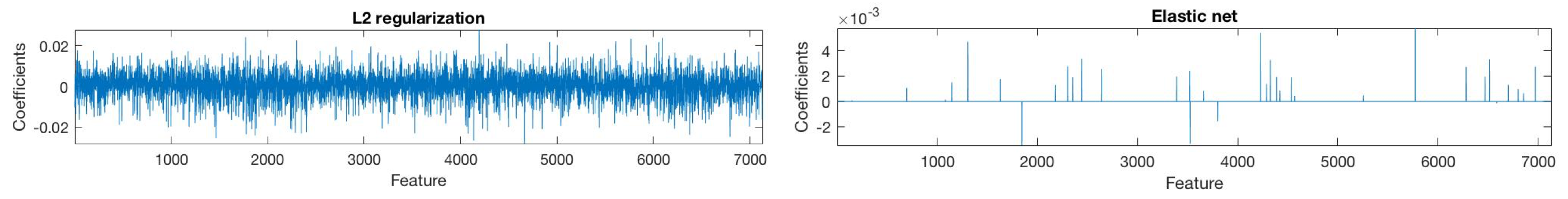}
    \caption{Coefficients for an L2 SVM (left) and elastic net SVM (right) trained on a gene expression microarray containing 7129 genes from 38 patients with either acute myeloid leukemia or acute lymphoblastic leukemia.}
    \label{fig:geneticdataset}
\end{figure}

In addition to this, we tested our SVM regularization approaches on an oligonucleotide array dataset containing probes for 7129 genes for 38 bone marrow samples\footnote{\url{http://mldata.org/repository/data/viewslug/leukemia-all-vs-aml}}: 27 acute lymphoblastic leukemia, 11 acute myeloid leukemia. Figure \ref{fig:geneticdataset} shows the resulting coefficients for the L2 SVM and elastic net SVM, and demonstrates the better interpretability of the sparse coefficient selection of the elastic net: by obtaining sparse coefficients, we can examine which genes are relevant for the classification task. In the case of the L2 regularization, this is not possible.


\begin{table} 
  \centering \small
  \begin{tabular}{ | r || r |r || r | r | r| r| }
  \hline
   & \multicolumn{2}{ |c|| }{Diagnostic breast cancer} &  \multicolumn{4}{ |c| }{Diagnostic breast cancer (contaminated)}  \\ \hline
  Method & Accuracy &  non-zero coeff. & Accuracy &  non-zero coeff. & \# relevant & \#  no relevant \\ \hline
  $L_2$-norm & 63.35(3.84) \%  & 12.2(1.99) & 63.48(3.08) \% &  31.7(2.62) & 12.2(1.98) & 19.5(2.17)  \\ 
  $L_1$-norm & 63.26(2.89) \%  & 12.5(2.12) & 63.10(4.25) \% &  30.5(2.54)  & 11.6(0.84) & 18.9(2.38) \\ 
  Elastic net & 83.07(11.23) \%  & 8.6(3.53) & 79.15(11.11)\% &  15.8(13.24) & 8.6(5.18) & 7.2(8.21) \\ 
  $k$-support & 93.57(0.95) \%  & 9.2(3.56) & 94.52(1.15) \% &  19(2) & 8.8(1.64) & 10.2(0.83) \\ \hline
  \end{tabular}
  \caption{SVM classification results for breast tumor diagnosis dataset containing 30 input features computed from digitalized images of needle aspirates, with binary labels malignant/benign. The dataset on the right has been contaminated with 30 additional features drawn from a $\mathcal{N}(0,1)$ distribution. The relevant coefficients indicate the number of non-zero coefficients that correspond to the original 30 features, whereas the non-relevant coefficients refer to the ones corresponding to the additional 30 noise features. All results show averages for 10 repetitions, together with the standard deviation.}
  \label{tab:breastcancer}
\end{table}

\paragraph{Grouping effect} To illustrate the concept of grouping effect, we generate synthetic data as follows. We generate 30 training instances in each of two classes, $y_+$ and $y_-$. Each instance $\bm{x}_i$ is a $p$-dimensional vector with $p=30$, such that for each class $\bm{x}_i$ has a normal distribution with mean either $\bm{\mu}_+$ or $\bm{\mu}_-$ (for classes $+$ and $-$ respectively) and covariance matrix $\bm{\Sigma}$:
\begin{equation}
\bm{\mu}_+ = (\underbrace{1,...,1}_{5},\underbrace{0,...,0}_{25})^\intercal 
\qquad
\bm{\mu}_- = (\underbrace{-1,...,-1}_{5},\underbrace{0,...,0}_{25})^\intercal 
\qquad
 \bm{\Sigma} = 
\begin{bmatrix}
    \bm{\Sigma}^*_{5 \times 5} & \bm{0}_{5 \times 25}\\
    \bm{0}_{25 \times 5} & \bm{I}_{25\times 25}
\end{bmatrix}
\end{equation}
where the diagonal elements of $\bm{\Sigma}^*_{5 \times 5}$ are all 1, and the off-diagonal elements are all equal to 0.8. Therefore, $\bm{x}_+ \sim \mathcal{N}(\bm{\mu}_+, \bm{\Sigma})$ and $\bm{x}_- \sim \mathcal{N}(\bm{\mu}_-, \bm{\Sigma})$. To generate such data, we first compute the Cholesky decomposition $\bm{\Sigma} = \bm{L}\bm{L}^\intercal$, where $\bm{L}$ is a lower triangular matrix. Then, we generate $\bm{u} \sim \mathcal{N}(\bm{0},I)$ by multiple separate calls to the scalar Gaussian generator. Finally, we compute $\bm{x}_+ = \bm{\mu}_+ + \bm{L}\bm{u}$ which has the desired distribution with mean $\bm{\mu}_+$ and covariance $\bm{\Sigma}$. Note that for both classes, elements $x_1,...,x_5$ are highly correlated, with $\rho=0.8$. 

\begin{figure}[h] 
	\centering
	\includegraphics[width=1\textwidth]{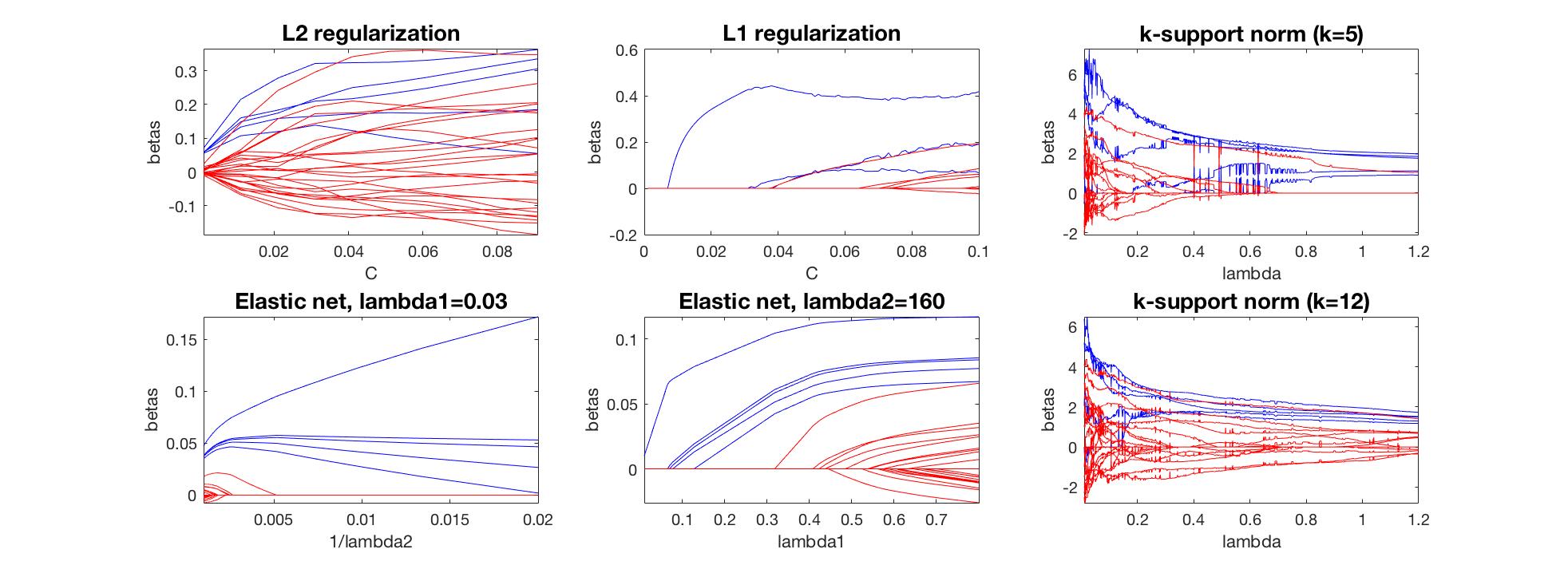}
    \caption{Comparison of the regularization paths of different SVMs on a simple synthetic dataset, where variables $x_1,...,x_5$ are highly correlated with $\rho=0.8$, whereas $x_6,...,x_{30}$ are not correlated. The blue lines indicate the coefficient values for the correlated variables, that is, $\beta_1,...,\beta_5$, whereas the read lines indicate the coefficient values $\beta_6,...,\beta_{30}$.}
    \label{fig:regularizationopaths}
\end{figure}

We trained different SVMs ($L_2$, $L_1$, elastic net, $k$-support) on this training set, and examined the resulting coefficients $\bm{\beta}$ for different combinations of hyperparameters. The results, shown in Figure \ref{fig:regularizationopaths}, indicate that the $L_2$-norm kept all variables in the model, the $L_1$-norm did variable selection but failed to identify the group of correlated variables, and the elastic net SVM successfully selected all five relevant variables, and shrunk their coefficients close to each other as expected (see Section  \ref{sec:elasticnet}), satisfying Equation  \ref{eq:groupingeq2}, which states  $
|\hat{\beta}_j-\hat{\beta}_l| \le \frac{\sqrt{n} }{\lambda_2} \sqrt{2(1-\rho)}
$ for two correlated variables $\bm{x}_j,\bm{x}_l$. The $k$-support SVM also performed grouped selection, shrinking the relevant coefficients close to each other, for large values of $\lambda$.

\subsection{L1 multi-class SVM vs one-versus-all SVM}

\paragraph{Synthetic dataset} To examine the performance of our multi-class $L_1$-regularized SVM (section \ref{sec:l1lasso}), we consider a four category classification problem. First, we sample 100 instances $\bm{u}_i = (u_{i,1},...,u_{i,100})$ from $\mathcal{N}(0,1)$. Then, we randomly assign them to four classes $c=1,...,4$, with 25 instances in each class. Finally, a linear transformation is performed: $x_{i,j} = x_{i,j} + a_j$; $j=1,2 $ and $x_{i,j} = x_{i,j}$; $j=3,...,100 $, with $a_j=(d,0),(0,d),(-d,0),(0,-d)$ for classes 1-4, respectively, for different values of $d$. In this example, only two features $x_j; j=1,2$ are relevant to classification, whereas the remaining 98 features are redundant. The hyperparameters of the different SVM algorithms tested (see Table \ref{tab:multiclass}) are tuned over a discrete set in $[10^-4,10^4]$ using leave-one-out cross-validation. The accuracy of the algorithms are examined in a different test set, constructed in the same way as the training set. Their optimal test errors, as well as the number of selected features (by both the binary and the final multi-class classifiers) were averaged over 50 repetitions and reported in Table \ref{tab:multiclass}. 

The results show that L1MSVM outperforms all algorithms in terms of number of number of features selected for all values of $d$, with the number of non-zero coefficients close to 2 (the number of relevant features in our dataset). In addition to this, L1MSVM also resulted in low classification errors (although it is outperformed by the OVA k-support for $d=2$ and by the OVA elastic net for $d=1,3$).

\paragraph{Gene expression dataset} Our $L_1$ multi-class SVM algorithm was also tested on a real dataset\footnote{\url{https://github.com/ramhiser/datamicroarray/wiki/Golub-(1999)}} containing gene expression data for 742 genes from 72 patients with leukemia. The training data consisted of 38 samples in three classes: 25 samples of acute myeloid leukemia (AML), 38 samples of B-cell acute lymphoblastic leukemia (B-ALL) and 9 samples of T-cell acute lymphoblastic leukemia (T-ALL). The test set contained 34 samples, with 14 AML, 19 B-ALL and 1 T-ALL. We tuned the algorithm hyperparametrs using leave-one-out cross-validation. Our results showed that our $L_1$ multi-class SVM algorithm  lead to a test error of 6\% and 4 non-zero coefficients. In contrast, the OVA methods select a large number of non-zero coefficients. For example, although the OVA elastic net resulted in a test error of 6\% as well, it   produced 77 non-zero coefficients. Therefore, the results provided by the L1MSVM offer better interpretability and. interestingly, all four genes selected by the L1MSVM have been individually or jointly identified as one of the predictor genes to T-ALL, B-ALL and AML\footnote{\texttt{http://www.leukemia-gene-atlas.org/LGAtlas/LGAtlas.html}}.

\begin{table} 
  \centering \small
  \begin{tabular}{ c l r r r}
  \hline
  Distance & Algorithm & Test error & \# features (binary) & \# features (multi-SVM) \\ \hline
  d=1 & OVA $L_2$ 			& 64.37(12.92) \% & 6.30(6.90) & 9.40(9.22) \\ 
      & OVA $L_1$ 			& 46.10(5.67) \% & 7.88(4.39) & 15.15(4.92) \\ 
      & OVA elastic net 	& 43.90(3.40) \% & 3.76(5.21) & 4.50(5.52) \\
      & OVA k-support 		& 45.60(5.36) \% & 6.65(4.53) & 12.90(6.38) \\
      & L1MSVM			 	& 44.57(4.04) \% & 2.77(2.66) & 4.31(2.71) \\ \hline
  d=2 & OVA $L_2$ 			& 38.03 (22.44) \% & 9.04(7.36) & 12.40(9.18) \\
      & OVA $L_1$ 			& 17.20(0.06) \% & 3.79(3.91) & 7.95(5.60) \\
      & OVA elastic net 	& 17.05(4.24) \% & 3.35(4.50) &  4.50(4.77)\\
      & OVA k-support 		& 14.50(3.92) \% & 4.88(3.31) &  10.30(5.58) \\
      & L1MSVM			 	& 15.85(3.01)\% & 2.5(0.51) & 4(1.1)\\\hline
  d=3 & OVA $L_2$ 			& 26.15(27.47) \% & 9.30(6.45)  & 13.73(8.54f) \\
      & OVA $L_1$ 			& 4.05(2.59) \% & 2.31(2.29) &  5.60(5.28) \\
      & OVA elastic net 	&  3.30(1.80) \%   & 3.36(4.80)  & 4.30(4.89) \\
      & OVA k-support 		& 5.20(2.15) \% & 4.30(2.31) &  10.10(5.38) \\
      & L1MSVM			 	& 3.71(1.38) \% & 2.18(0.15)   &  2.44(0.41)    \\\hline
  \end{tabular}
  \caption{Average test errors,  number of features selected by the binary classifier and number features selected by the multi-class classifiers\protect\footnotemark, together its standard errors (in parenthesis), over 100 simulation replications.}
  \label{tab:multiclass}
\end{table}
\footnotetext{We consider that a feature has been selected by a multi-class classifier if it has been selected by any of the corresponding binary classifiers.}





\section{Conclusion}

In this project, we have examined a variety of regularization approaches for binary and multi-class classification with support vector machines. The sparsity properties of these regularization approaches were demonstrated in both synthetic and real biomedical datasets, showing how more sparse solutions can lead to better interpretability of the learning results. Furthermore, we showed mathematically  the grouping property of the elastic net (not present in L1 and L2 regularization) and demonstrated it empirically on a synthetic dataset. Unfortunately, due to the limited scope of this project, we couldn't explore other interesting regularization approaches that could offer interesting features for biomedical data classification, such as group lasso (recently used in magnetic resonance imaging data classification) or  sparse multiple kernel learning with SVMs. These will remain topics for future exploration.


\nocite{*}  
{\small
\bibliography{biblio}{}
\bibliographystyle{plain}
}


\end{document}